\documentclass[a4paper,fleqn]{cas-dc}

\usepackage{soul}
\usepackage{color, xcolor} 
\usepackage[numbers]{natbib}
\usepackage{tabularx}

\begin{document}

\shorttitle{Large Language Models in Law: A Survey} 
\shortauthors{Jinqi Lai \textit{et al.}}


\title [mode = title]{Large Language Models in Law: A Survey}

\author[1]{Jinqi Lai}
\ead{jqlai499@gmail.com}
\address[1]{Jinan University, Guangzhou 510632, China}  

\author[1]{Wensheng Gan}
\cortext[cor1]{Corresponding author}
\ead{wsgan001@gmail.com}
\cormark[1]

\author[1]{Jiayang Wu}
\ead{csjywu1@gmail.com}

\author[2]{Zhenlian Qi}
\ead{qzlhit@gmail.com}
\address[2]{Guangdong Eco-Engineering Polytechnic, Guangzhou 510520, China}

\author[3]{Philip S. Yu}
\ead{psyu@uic.edu}
\address[3]{University of Illinois Chicago, Chicago, USA}

\begin{abstract}
  The advent of artificial intelligence (AI) has significantly impacted the traditional judicial industry. Moreover, recently, with the development of AI-generated content (AIGC), AI and law have found applications in various domains, including image recognition, automatic text generation, and interactive chat. With the rapid emergence and growing popularity of large models, it is evident that AI will drive transformation in the traditional judicial industry. However, the application of legal large language models (LLMs) is still in its nascent stage. Several challenges need to be addressed. In this paper, we aim to provide a comprehensive survey of legal LLMs. We not only conduct an extensive survey of LLMs, but also expose their applications in the judicial system. We first provide an overview of AI technologies in the legal field and showcase the recent research in LLMs. Then, we discuss the practical implementation presented by legal LLMs, such as providing legal advice to users and assisting judges during trials. In addition, we explore the limitations of legal LLMs, including data, algorithms, and judicial practice. Finally, we summarize practical recommendations and propose future development directions to address these challenges.
\end{abstract}


\maketitle

\section{Introduction}  \label{sec:Introduction}

Artificial intelligence (AI) \cite{brynjolfsson2017artificial,fetzer1990minds,zhang2021study}, a term that first emerged in the 1950s, was initially defined as ``the science and engineering of making intelligent machines". AI integrates various fields of knowledge, such as computer science, psychology, and mathematical physics. Its goal is to enable machines to understand data and make decisions akin to human intelligence. Over the past 60 years, AI has undergone development and iteration, leading to more refined definitions. In its early stages, AI gained attention with the ``Turing Test" and the chess-playing program. With the advancement of machine learning and the introduction of AI learning frameworks, AI technologies started gaining prominence and finding applications in various domains. In 2006, deep learning \cite{goodfellow2016deep,lecun2015deep} was formally proposed by Hinton. With the continuous development of technologies such as big data \cite{shorten2019survey}, the Internet of Things (IoTs) \cite{laghari2021review} and AI, deep learning has made significant progress. For example, the IBM question-and-answer robot is widely used in the field of speech recognition \cite{saon2015ibm}, as well as the emergence of driverless technology. In addition, AlphaGo can compete with the world's top Go players \cite{silver2017mastering}. AI-generated content (AIGC) \cite{cao2023comprehensive,wu2023ai} is a new AI technology that follows in the footsteps of professional-generated content (PGC) and user-generated content (UGC) \cite{tu2021ugc}. AIGC encompasses various techniques, such as generative adversarial networks (GANs) \cite{pan2019recent} and diffusion models \cite{croitoru2023diffusion}. AIGC enables the generation of relevant content based on given instructions. It has great potential applications in the fields of image resolution \cite{zhang2019graph} and audio recognition \cite{shirian2021compact}.

Foundation models \cite{bommasani2021opportunities}, also known as large models, represent an important direction in the development of AI technologies and were introduced in 2021. The Seq2Seq large model \cite{keneshloo2019deep}, based on sequence modeling, includes the Encoder-Decoder framework \cite{park2018sequence}, which encodes input sequences and decodes them into the desired output sequences. Large models are used in text transformation tasks such as machine translation \cite{sutskever2014sequence}, interactive chat responses \cite{zhang2019dialogpt}, and content generation \cite{liu2021deep}. The text transformation tasks could be divided into two main domains. In the audio-to-text domain, large models can convert input text into speech sequences and perform speech recognition \cite{prabhavalkar2017comparison}. In the image-to-text domain, large models can generate images based on user-provided descriptions \cite{venugopalan2015sequence}. However, large models face challenges when dealing with long input sequences due to limitations in vector sizes, which can result in issues like blurriness when compressing large images. Attention mechanisms \cite{niu2021review} were introduced to address this problem. They assign weights to important information by processing input queries, keys, and values. The query represents the user's input question, and the attention mechanism converts the query into attention values that are most relevant to the input query. By calculating the similarity between the query and key and aggregating them with the weighted values, attention mechanisms can enhance the performance of large models \cite{niu2021review}. The Transformer framework \cite{vaswani2017attention}, proposed by Vaswani in 2017, laid the foundation for large model architectures. In addition, Google introduced the BERT model \cite{kenton2019bert} for pre-training, and OpenAI developed GPT \cite{radford2018improving}. Then, numerous pre-training language models such as ELNet and RoBERTa \cite{liu2019roberta}, and GPT-2 \cite{radford2019language} emerged. High-parameter models like Megatron-LM \cite{shoeybi2019megatron}, T5 \cite{raffel2020exploring}, and Turing-NLG were proposed. In 2020, GPT-3 \cite{brown2020language} was introduced, capable of generating coherent paragraphs of text. It is composed of the MT-NLG \cite{smith2022using} natural language generation model, which consists of 5.3 trillion parameters. In 2021, the first trillion-parameter language model, Switch Transformer \cite{fedus2022switch}, was established. In addition, the universal sparse language models like GLaM \cite{du2022glam}, INTERN large models \cite{shao2021intern}, PanGu-$\alpha$ NLP models \cite{zeng2021pangu}, and Chinese pre-training language models like PLUG \footnote{https://m.thepaper.cn/baijiahao\_12274410} were established. In 2022, Stability AI released the Diffusion model for text-to-image generation \cite{ruiz2023dreambooth}. ChatGPT \cite{radford2018improving} was trained by using supercomputers and code. The ``brain-level AI model" BAGUALU was released \cite{ma2022bagualu}. These large models are extensively applied in fields such as education \cite{chen2020artificial,gan2023large,kasneci2023chatgpt}, medicine \cite{hamet2017artificial}, robotics \cite{zeng2023large},  and others \cite{gan2023model}.

In the long history of law and justice, courtroom rulings with judges taking the lead, and the balance of power between prosecution and defense have always been crucial factors in resolving legal cases. The law possesses authority, rigor, objectivity, and normativity, while the judicial process is characterized by periodicity and fairness. However, with the increasing population, the number of judicial cases has also grown, and the imbalance between the number of cases and the available human resources has led to prolonged judicial processes. Relying solely on human judgment is no longer sufficient to meet societal demands. Therefore, the application of artificial intelligence in the field of law has become significant \cite{rissland2003ai}. In recent years, with the continuous development of deep learning and other AI technologies, the legal field has witnessed the emergence of more intelligent applications \cite{chen2019deep}. For example, some regions have introduced the concept of smart courts, or AI courts, to assist in the adjudication process using AI technology. Examples of legal applications of AI include legal research, document analysis, contract review, predictive analytics for case outcomes, and legal chatbots for providing basic legal information and guidance. AI can help improve efficiency, accuracy, and accessibility in the legal system, but it is important to ensure transparency, accountability, and ethical considerations when implementing AI in legal processes.

\textbf{Research gaps}. The law and AI applications have widely adopted many related AI technologies \cite{atkinson2020explanation,rissland2003ai,sourdin2018judge,surden2019artificial}. The rise of LLMs has brought more possibilities for assisting law management and decision-making. This paper is the first comprehensive review of legal LLMs, introducing the definition, significance, applications, and future. In our paper, we mainly address the following three questions: What is the future of the law and AI? What are the characteristics of legal LLMs, and what are their shortcomings? Will it be possible to replace the judges' role in the legal system? Furthermore, we propose some suggestions for improving legal LLMs.

\textbf{Contributions}. To fill in the gaps in current research, we first conducted a systematic literature review on the use of LLMs in the judicial field. What's more, we also show our opinions on the future legal LLMs system. The main contributions of this paper are as follows:

\begin{itemize}
    \item To the best of our knowledge, this is the first review article on legal LLMs. We first explain the basic concepts related to AI and law. Following that, we provide a detailed description of the characteristics of legal LLMs, showcasing the difference between traditional judgement and current judgement with AI.
    
    \item We demonstrate how judges can make fairer decisions with the help of legal LLMs. We also explore in detail how to promote the optimization of legal LLMs based on the characteristics of big legal data and the intricacies of judicial practice.
    
    \item We provide the latest research on legal LLMs, including studies conducted by companies and universities, as well as an in-depth investigation of fine-tuning model techniques and evaluation strategies.
    
    \item  Finally, we summarize the key challenges and future directions of legal LLMs and provide suggestions and directions for improvement.
\end{itemize}

\begin{figure}[ht]
    \centering
    \includegraphics[scale=0.38]{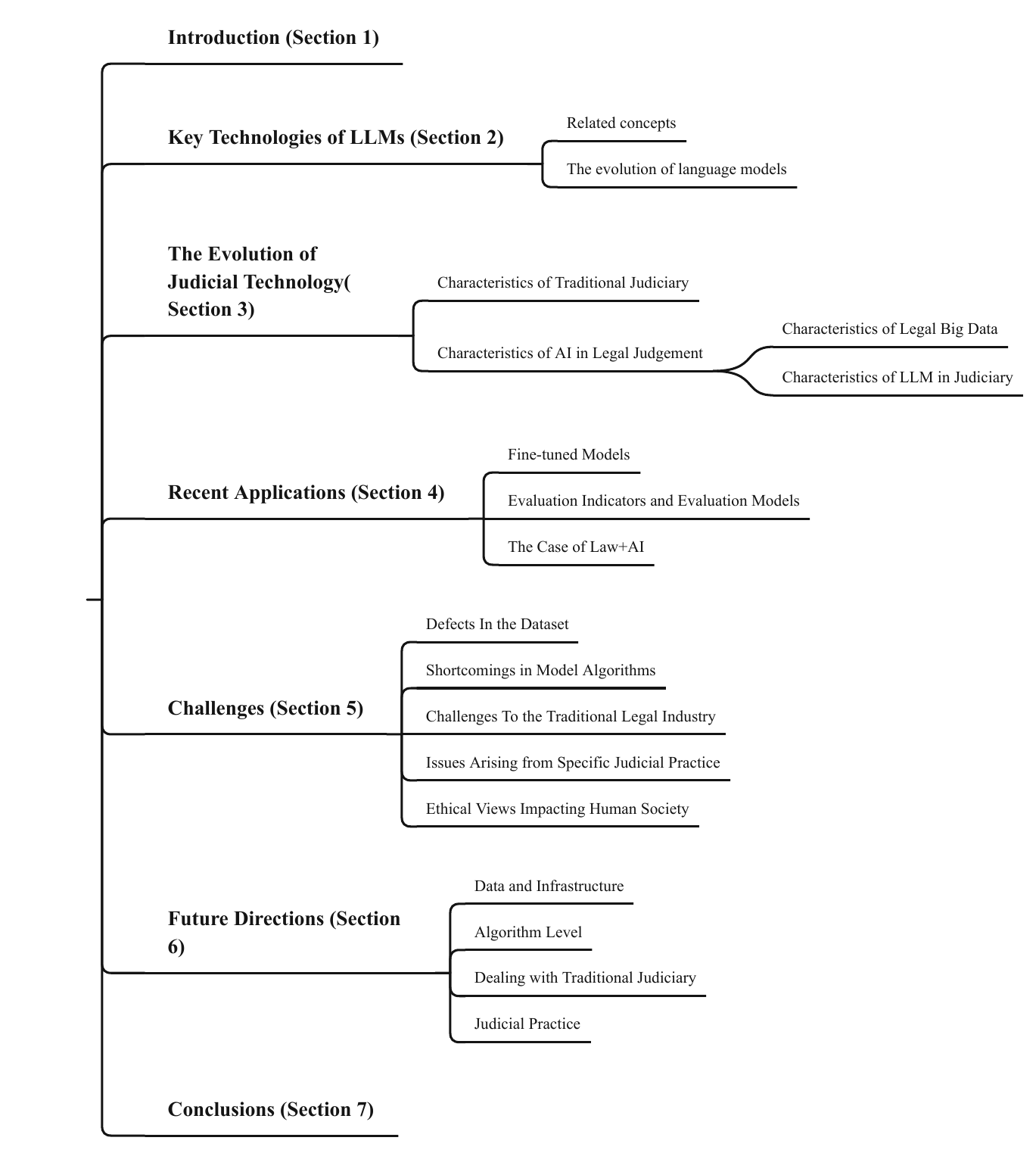}
    \caption{The outline of our overview.}
    \label{fig:outline}
\end{figure}

\textbf{Organization.} The organization of this paper is shown in Fig. \ref{fig:outline}. In Section \ref{sec:key}, we introduced the key technologies and development history related to LLMs. We provide a brief overview of the traditional judiciary, legal big data, and the main characteristics of AI judiciary in Section \ref{sec:evolution}. In Section \ref{sec:applications}, we based on the recent popular fine-tuned legal large models and model evaluation methods, focusing on their applications. In Section \ref{sec:challenges}, we analyzed the challenges faced by legal LLMs from the perspectives of legal data, algorithms, traditional judiciary, judicial practice, and human social ethics. Then, we provided specific feasible directions for future research in Section \ref{sec:future}. Finally, we gave the conclusion of this survey in Section \ref{sec:conclusions}.

\section{Key Technologies of LLMs} \label{sec:key}
\subsection{Related Concepts}

In this subsection, we explore several related concepts that are fundamental to understanding the field of AI. These concepts encompass various aspects of AI, machine learning, neural networks, deep learning, and natural language processing (NLP). Understanding these concepts is crucial for grasping the underlying principles and techniques behind LLMs and their role in advancing NLP tasks.

\textbf{Artificial intelligence (AI)} \cite{russell2010artificial} refers to the emulation of human intelligence, thinking processes, and behavior in machines. The objective of AI is to enable computers to possess abilities such as understanding, learning, and reasoning. AI achieves these capabilities primarily through data acquisition and processing. With these capabilities, AI has found wide-ranging applications across various industries. For example, in the field of healthcare \cite{hamet2017artificial}, Google DeepMind's AlphaFold has been utilized to predict protein structures, assisting doctors in diagnosing and treating diseases. In the fields of finance \cite{cao2022ai}, companies like BlackRock employ AI technology for quantitative investment, aiming to enhance investment returns and aid in decision-making processes. Furthermore, in the fields of agriculture \cite{eli2019applications}, AI has led to the development of smart farming machinery for precise seeding and fertilization, optimizing crop yields.

\textbf{Machine learning (ML)} \cite{jordan2015machine} primarily focuses on studying how machines can process data to acquire new skills or perform tasks such as prediction and decision-making by simulating or implementing human learning behaviors. It can be described as searching for the relationship between the feature dataset $X$ and the label dataset $Y$. ML can be categorized into supervised learning, unsupervised learning, and semi-supervised learning. 

\textbf{Neural network (NN)} \cite{gurney2018introduction} is a branch of machine learning and a core component of deep learning. It simulates the neural networks of the human brain by connecting multiple ``neurons" through connectionism to process information. Each neuron receives input signals, weights and activates them, and passes the results to other neurons. Neural networks can be classified into fully connected neural networks, convolutional neural networks, recurrent neural networks, etc.

\textbf{Deep learning (DL)} \cite{lecun2015deep} is a type of machine learning algorithm. The term ``deep" implies the use of multiple layers in neural networks, enabling computers to learn from a large amount of data and extract useful features and patterns. With multi-layered neural networks, deep learning can handle large-scale data and perform deeper feature learning, classification, and prediction. Typical deep learning frameworks include PyTorch \cite{paszke2019pytorch}, TensorFlow \cite{abadi2016tensorflow}, MindSpore \cite{han2021transformer}, and Caffe \cite{jia2014caffe}. 

\textbf{Self-attention mechanism} \cite{shaw2018self} is a crucial technique in deep learning and has been extensively used in processing sequential data in NLP. By encoding the input sequence into different vectors in the embedding layer, the self-attention mechanism dynamically assigns and calculates attention weights for different parts and outputs the weighted sum as the sequence representation. The self-attention mechanism often employs a multi-head mechanism to capture different relationships and features, including long-range dependencies. It finds wide application in the NLP models. For example, the Transformer model incorporates a self-attention mechanism to achieve optimization and breakthroughs \cite{yang2022transformers}.

\textbf{Natural language processing (NLP)} \cite{chowdhary2020natural,nadkarni2011natural} serves as a crucial bridge between machines and human natural language. Its primary goal is to enable machines to understand, generate, and manipulate human language, facilitating seamless communication between humans and machines. NLP has a wide range of applications across various industries. One prominent application is as the role of search engines \cite{cafarella2005search}, where NLP techniques are used to understand user queries and retrieve relevant information from vast amounts of text data. In the legal domain, NLP is utilized as a tool for information extraction from legal documents. It could help to extract specific legal elements, such as case references or contract clauses \cite{sleimi2018automated}, from large volumes of legal text. Furthermore, NLP technology is an essential component of chatbots \cite{regin2022automated} and virtual assistants \cite{campagna2019genie}. It enables these systems to generate human-like responses based on user input, facilitating interactive and engaging conversations.

\textbf{Multi-task learning (MTL)} \cite{zhang2021survey} can improve model performance by  setting multiple tasks. In MTL, the model is designed to share underlying representations to handle multiple tasks instead of training multiple independent models for each task. MTL exhibits generalization capabilities that extend to unknown datasets. Moreover, by learning across multiple tasks, MTL enhances efficiency and reduces resource consumption. 

\textbf{Foundation model}: The concept of a foundation model, proposed by OpenAI, refers to a large-scale pre-training model \cite{bommasani2021opportunities}. These models are trained on massive amounts of data and can be fine-tuned for various tasks. They provide a solid ``foundation" for various applications. Foundation models are powerful models obtained through pre-training on large-scale data. They serve as the basis for constructing various specific AI applications, providing a reliable starting point for multiple tasks. These foundation models employ various AI techniques and have achieved remarkable performance in natural language understanding and generation.

\textbf{LLMs}: A LLM \cite{wei2022emergent,zhao2023survey} refers to a language model with a complex structure and a large number of parameters. LLMs are typically trained on vast amounts of unlabeled data through a pre-training stage \cite{carlini2021extracting}. The step of pre-training equips LLMs with the capability to perform various NLP tasks \cite{min2023recent}, including automatic text generation and translation. What's more, the core technologies utilized in LLMs include fine-tuning and reward modeling. These techniques contribute to the development and refinement of LLMs, enhancing their language understanding and generation capabilities. Nowadays, LLMs have been widely used in different natural language tasks. They are employed in intelligent translation systems \cite{brants2007large} enabling accurate translation, such as Google's neural machine translation (NMT) \cite{wu2016google}. LLMs are also utilized in text generation tasks, where models can produce poems or texts in specific styles based on prompt words. Furthermore, LLMs play a crucial role in intelligent customer service and question-answering systems, such as virtual assistants Apple's Siri \cite{kepuska2018next} and Amazon's Alexa \cite{hoy2018alexa}. These systems could help to understand and respond to user inquiries in a conversational manner. Moreover, LLMs have contributed to advancements in speech recognition like Microsoft's Cortana \cite{hoy2018alexa} and Google's speech recognition system \cite{kepuska2018next}. They could help convert spoken language into written text.

\subsection{Evolution of LLMs}

\begin{figure}[ht]
    \centering
    \includegraphics[scale=0.3]{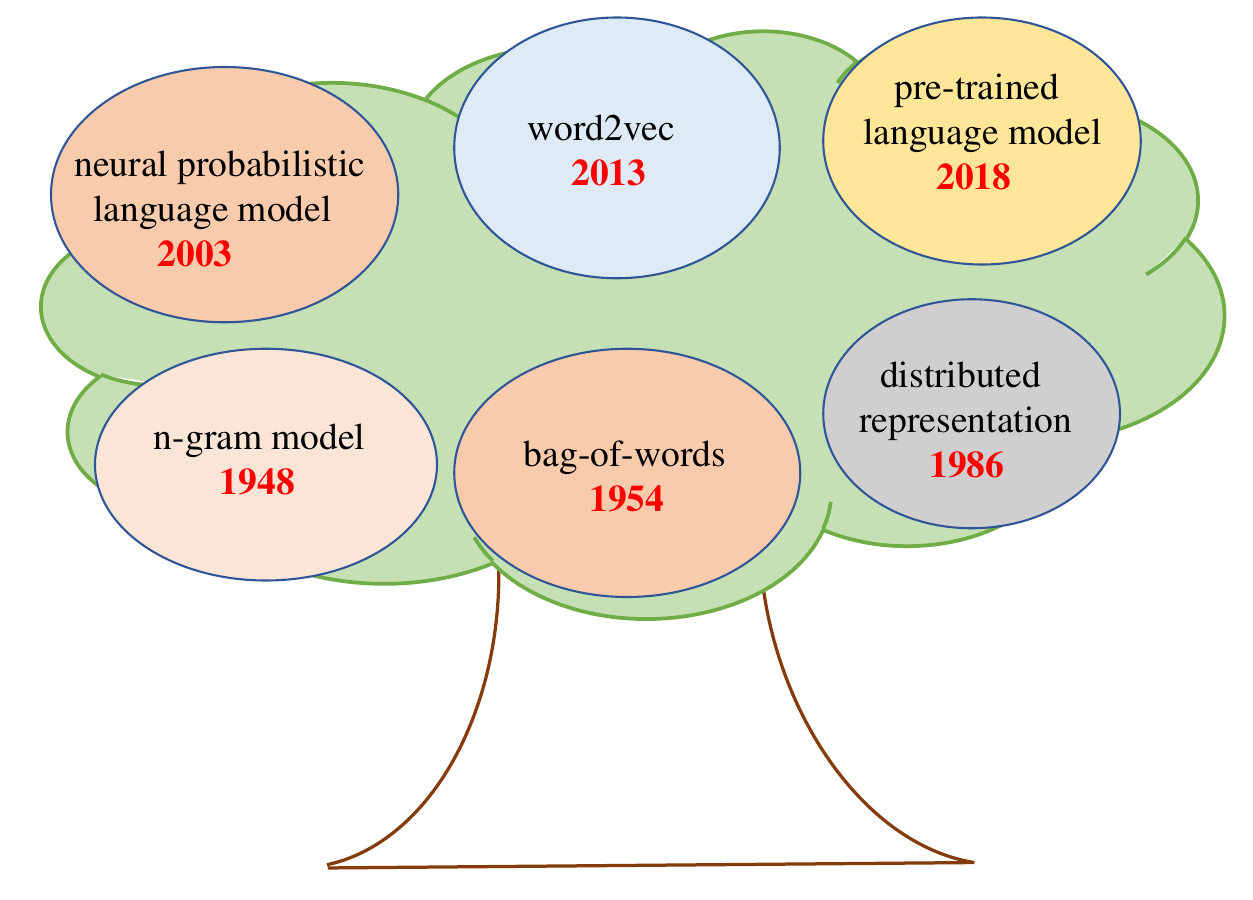}
    \caption{The evolution of language models.}
    \label{fig:evolution}
\end{figure}

The evolution of language models can be described through six periods, as shown in Fig. \ref{fig:evolution}. As early as 1948, the emergence of N-gram models \cite{brown1992class} divided the text into combinations of n-words and used statistical methods to predict the probability of the next word. These models were unable to capture complex language dependencies and semantic structural information. Then, in 1954, the bag-of-words model \cite{zhang2010understanding} appeared, treating text as a collection of words but disregarding word order and semantic relationships. The limitations of these models led researchers to seek better methods for understanding and generating natural language. In 1986, the concept of distributed representations \cite{hinton1986learning} was introduced, considering the semantic and contextual information of vocabulary, which brought new ideas for the development of language models.

In 2003, neural probabilistic language models \cite{bengio2000neural} introduced neural networks in an attempt to improve model performance through the learning capacity of neural networks. However, due to limitations in the scale of neural networks and training data, the processing capability of these models remained limited. Subsequently, in 2013, the emergence of the Word2Vec model \cite{church2017word2vec} provided a new way to represent the language model's capabilities by training and learning word vectors to capture semantic relationships between words. The most significant breakthrough occurred in 2018 when pre-training language models such as BERT \cite{kenton2019bert} and GPT \cite{radford2018improving} were introduced. These models, based on the Transformer architecture, were pre-trained on large-scale text data and learned rich semantic and syntactic knowledge \cite{voita2019bottom}. They could model contextual information in text in a bidirectional or generative manner, possessing stronger language understanding and generation capabilities. In 2020, the release of the T5 model transformed various NLP tasks into text-to-text transformation problems and achieved excellent performance \cite{ni2021sentence}. Besides, GPT-3 became one of the largest pre-training language models \cite{zeng2023distributed}, showcasing powerful generation capabilities. Moreover, between 2022 and 2023, models such as GPT-3.5\footnote{https://en.wikipedia.org/wiki/GPT-3} and GPT-4\footnote{https://openai.com/gpt-4} further improved and enhanced the scale, performance, and generation capabilities. The consistent evolution of these language models reflects development and technological advancement. Each improvement has enhanced the model's expressive and processing capabilities, providing more opportunities and possibilities for the application of AI in the legal domain.

\section{Evolution of Judicial Technology} \label{sec:evolution}

In this section, we explore the development of judicial AI. By analyzing the shortcomings of traditional judiciary and the characteristics of legal big data, we further uncover the important features of the LLM judiciary and discuss them in depth.

\subsection{Characteristics of Traditional Judiciary}

We explore some characteristics of the traditional judiciary that are crucial for our subsequent understanding of the application of judicial artificial intelligence  \cite{re2019developing}. These characteristics include reliance on human decision-making, lack of flexibility, and resource consumption, among others. They are outlined as follows:

\textbf{Reliance on human decision-making}: Traditional judicial systems primarily rely on the human decision-making of judges, prosecutors, and lawyers, including case hearings, judgments, and legal interpretations. During the process of reasoning and evidence collection in a case, they often need to refer to the specific circumstances of the case, legal provisions, and past precedents, combined with their professional knowledge, to make judgments and decisions. Finally, the judgment or defense is conducted through a trial.

\textbf{Precedent-based}: Traditional judiciary often relies on precedents in the decision-making process \cite{baude2020precedent}, such as previous judgments in similar cases and relevant provisions of laws. In many judicial systems, the judgments of the highest court have authority and binding effect, guiding other courts. For example, the decisions of the highest court are considered authoritative for constitutional interpretation and legal application, and other courts often refer to those decisions in relevant cases.

\textbf{Lack of flexibility}: In the field of judiciary, when there is uncertainty within the legal norms or vague boundaries of legal concepts, judges need to make judgments and choices based on specific contexts rather than mechanically applying the law \cite{easterbrook1987stability}. For example, when dealing with contracts, if a clause involves the concept of ``reasonable time", but the contract does not provide a specific definition, the judge needs to determine what constitutes a ``reasonable time" according to the specific circumstances. In this case, the judge should consider the nature of the contract, the relationship between the parties, industry standards, and other relevant factors. Therefore, judges have the characteristic of flexibility in handling judicial cases.

\textbf{Time and resource-consuming}: Traditional judiciary requires a significant amount of human resources and time when dealing with a large number of cases. This leads to situations where there are many cases but few personnel, and it can prolong the trial process. For example, the process of case hearings, summoning witnesses, and collecting evidence may consume a considerable amount of time and resources. Similar hierarchical systems also exist in other countries as well. For instance, the judicial system in the United Kingdom adopts a hierarchical system \cite{hanretty2020court}, including the Magistrates' Court, the Crown Court, and the Supreme Court.

\subsection{Characteristics of AI in Legal Judgement}

Legal professionals can utilize the logical reasoning capabilities of legal LLMs to understand the process of cases, assist judges in decision-making, quickly identify similar cases through language comprehension, analyze and summarize key case details, and use automated content generation capabilities to draft repetitive legal documents. By alleviating the issue of  ``too many cases, too few people", these AI systems can enhance judicial efficiency and quality \cite{re2019developing}. There are some characteristics of legal big data and LLMs shown as follows:

\subsubsection{Characteristics of Legal Big Data}

Processing data sets is an important part of training a large model. Legal big data, compared to other datasets, exhibits non-structured, multi-sourced, timely, and privacy and security features, among others, which have garnered attention. \textbf{Unstructured}: Legal data often exhibits unstructured characteristics, such as legal concepts, legislative texts, judgments, legal commentaries, etc. The textual data formats are inconsistent and not easily understood and processed by computers \cite{adnan2019analytical}. Therefore, NLP and text analysis techniques are required to extract useful information and transform it into structured data that can be understood by AI.

\textbf{Multilingual and multicultural}: Law covers multiple languages and cultures \cite{vsarvcevic2016basic}, so legal big data may involve texts in different languages. Cross-lingual analysis of legal data requires addressing challenges such as translation, cultural differences, and variations in legal terminology. For example, European Union regulations often have multiple official language versions, such as English, French, German, etc. Legal researchers need to compare different language versions of regulations to ensure an accurate understanding of their meanings.

\textbf{Vast scale and complexity}: Legal data usually contains a large amount of text, such as hundreds of pages of legislation or judgments. Moreover, legal data exhibits diverse characteristics, with each domain having its own unique regulations and types of legal documents, requiring different analysis methods and specialized knowledge. For instance, the field of intellectual property involves various legal issues, including patents, trademarks, copyrights, etc. These datasets are extensive and require powerful computing capabilities for processing and analysis. Some legal data, such as legal statutes, can be complex, and due to their large volume and diversity, AI needs robust computational power for handling and analysis. For example, tax law is highly intricate, encompassing thousands of pages of tax regulations and judicial precedents. This requires large-scale legal databases and advanced search tools to assist tax professionals in research and analysis \cite{zhong2020does}.

\textbf{Timeliness}: Law is an ever-changing field, with legal documents and regulations being frequently revised and updated. Therefore, legal big data must be regularly updated to reflect the latest legal provisions. For example, in tax law, the government may introduce new tax regulations to adapt to economic changes. Thus, legal professionals and tax experts need to keep their research and advice up-to-date.

\textbf{Data multi-sourcing}: Legal big data can come from multiple sources, including court records, government files, legal databases, and social media. Integrating data from these different sources and ensuring data consistency is a challenge. For example, legal researchers may need to access both federal and state court records to obtain comprehensive case information, which requires integrating data from different sources.

\textbf{Privacy and security concerns}: Legal data may contain labeled sensitive information, such as personal identifying information, details of privacy-related cases, etc. Therefore, during the collection, storage, and analysis of legal big data, operations like anonymization are needed to remove privacy labels. Moreover, protecting user privacy is crucial \cite{liu2021machine}. For example, court documents in criminal cases may contain personal identification information and the criminal histories of defendants. These pieces of information require strict privacy and security protection to prevent unauthorized access and disclosure.

\subsubsection{Characteristics of LLM in Judiciary}

With the rapid development of AI technologies such as AIGC, the application of legal LLMs in the field of judiciary has become widespread. They exhibit the following characteristics, as shown in Fig. \ref{fig:lawcharacteristic}:

\begin{figure}[ht]
    \centering
    \includegraphics[scale=0.28]{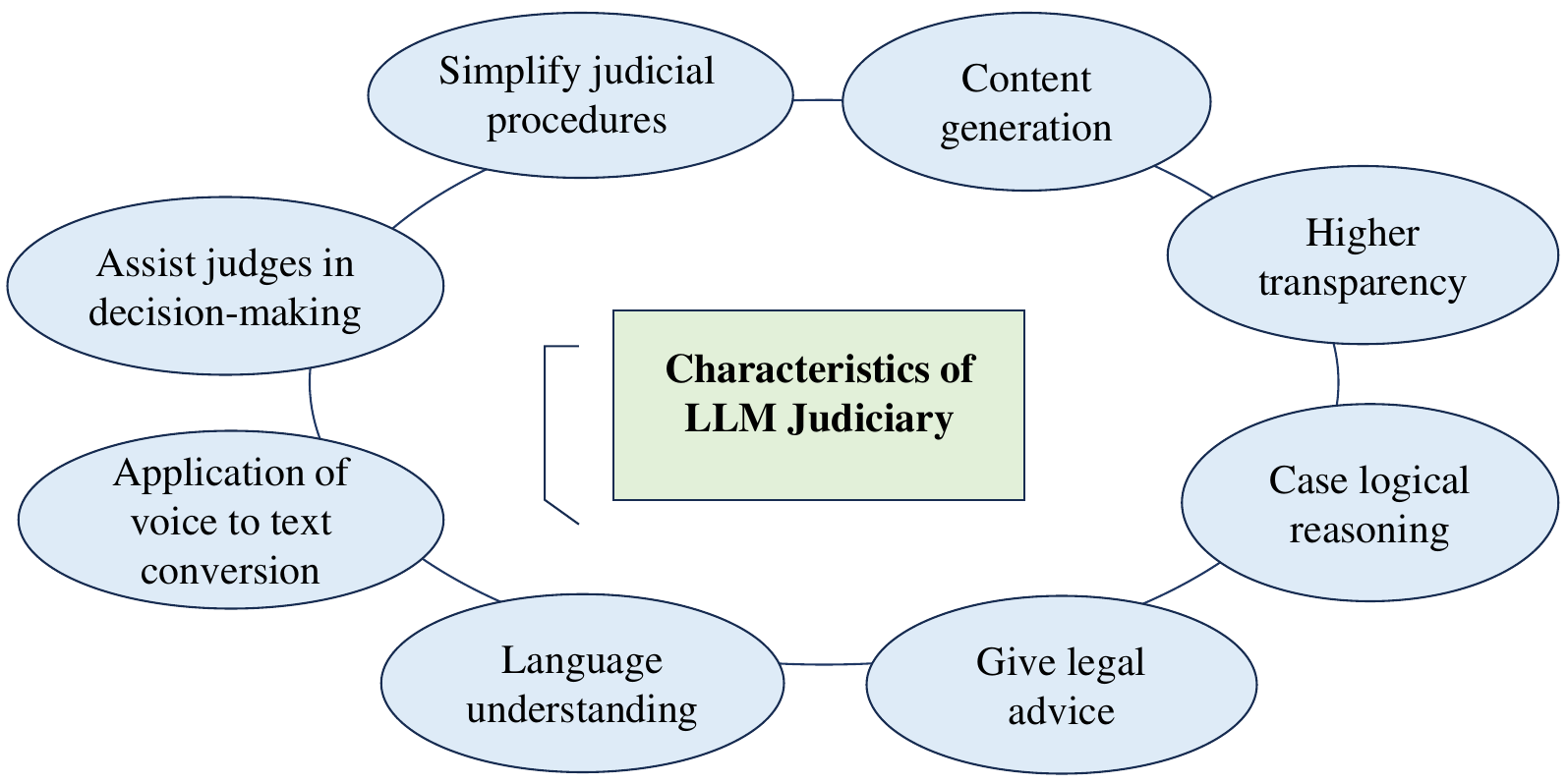}
    \caption{Characteristics of LLM in Judiciary.}
    \label{fig:lawcharacteristic}
\end{figure}

\textbf{Language understanding}: LLMs have the ability to interact with users and establish contextual relationships to perform various tasks \cite{du2022shortcut}. Through deep learning on extensive legal datasets, LLMs can analyze and modify legal documents, checking grammar, symbols, sentence structures, etc. They can also extract key elements from legal documents, such as the disputed issues in a case, applicable laws, the identities of the parties involved, the relevance of evidence, and more. These extracted features are crucial inputs for legal decision-making algorithms. LLMs can quickly extract key points from legal documents, combine them with the judgment outcomes, and generate concise and accurate case summaries. Legal professionals can utilize LLMs to extract key points from legal documents, combine them with judgment outcomes, and generate concise and accurate case summaries, reducing time and effort while still producing high-quality work.

\textbf{Content generation}: LLMs can automatically generate legal documents \cite{kanapala2019text}, case reports, legal contracts, etc. By inputting basic information about a legal case, such as party details, legal grounds, and evidence, AI algorithms can generate a draft legal document that complies with legal standards. For example, a lawyer can input the basic information of a legal case into an AI system, which will then generate an initial draft of a legal document containing relevant clauses, conditions, and vocabulary. The lawyer can review and modify the draft, saving time and effort while maintaining high-quality work. Moreover, AI technologies can also structure text, such as Markdown text, JSON text, Excel text, etc. Users can input legal report files and specify the desired text format, generating structured texts that meet their requirements.

\textbf{Application of speech-to-text}: In traditional court proceedings, judges need to handwrite meeting records and submit them to typists to complete information recordings. With the continuous development of AI technology, speech-to-text conversion \cite{reddy2013speech} has become widely used, reducing the incompleteness of manual records. AI-based speech-to-text platforms \cite{trivedi2018speech} also increase the transparency of AI involvement in decision-making. For example, in some foreign courts \cite{etulle2023investigating}, the process of speech-to-text conversion \cite{trivedi2018speech} is fully presented in the courtroom, ensuring the credibility of meeting records and increasing public trust in the judiciary. AI can also play an assisting role in judicial decision-making for judges \cite{xu2022human}.

\textbf{Provide legal consultation}: In the field of justice, LLMs have the ability to interact with users. Therefore, users can ask legal questions to the model and receive answers and suggestions based on the training data \cite{greenleaf2018building}. This approach can provide convenient and efficient legal consultation services for users, while also reducing the workload of professional lawyers. For example, in a civil lawsuit, the user can ask the model multiple questions and the model can provide better judgement plans based on legal knowledge \cite{zhong2020iteratively}.

\textbf{Matching optimal solutions for cases}: AI can extract the key features of a given case, deeply mine a large number of historical cases and judgment outcomes, and obtain the optimal solution for the case \cite{kallus2020generalized}. It can assist in comprehensive evidence-searching. Furthermore, judicial discretion models can capture the detailed points of a case based on its features, maximizing judicial fairness.

\textbf{Case logic reasoning}: Interactive AI can perform case logic reasoning to a certain extent based on multi-round prompts from users and given case-related information \cite{atkinson2020explanation}. It analyzes the evidence chain of a case by extracting relevant elements from input legal documents and mining internal details. Legal LLMs conduct comprehensive analysis based on metrics such as the relevance, credibility, validity, and completeness of evidence, thus establishing a complete evidence chain for the case. Similarly, AI evaluates the authenticity and detail of case facts by analyzing the input legal text information and the feasibility of the facts. For example, in some areas, expert inspections have revealed inconsistencies in case facts within the dataset used to train ChatGPT. Legal LLMs, based on deep learning from extensive datasets, possess a certain degree of decision-making capability. It can propose judicial decision suggestions based on facts and legal regulations to assist judges in decision-making. For example, it includes the implementation of ``smart courts" \cite{shi2021smart} and the popularity of legal LLMs \cite{zhong2020does}.

\textbf{Simplifying judicial procedures - improving judicial efficiency}: The ``too many cases, too few people" issue has long been a problem in the field of judiciary. Much of the work of legal professionals is based on repetitive document-based tasks. AI technologies, focused on how to incorporate more comprehensive legal big data into systems \cite{zhong2020does}, have simplified judicial procedures and to some extent improved judicial efficiency \cite{de2022artificial}. For example, AI applications in judicial scenarios have reduced judges' administrative work \cite{sourdin2018judge} and increased trial efficiency \cite{de2022artificial}.

\section{Recent applications} \label{sec:applications}

In this section, we combine the analysis of the latest popular ten fine-tuned legal LLMs to examine their distinct characteristics in handling legal matters. We then present feasible model evaluation metrics and methods and discuss some AI legal case studies.

\begin{table*}[ht]
    \setlength{\tabcolsep}{12pt} 
    \caption{The multimodal models}
    \label{table2}
    \begin{tabularx}{0.9\textwidth}{|m{2.1cm}<{\centering}|m{3cm}<{\centering}|X|}
    \hline
    \textbf{LLM-based law}  &  \textbf{Traditional model} & \textbf{Website}  \\
    \hline
    LawGPT$_{zh}$  & ChatGLM-6B LoRA 16bit & \url{https://github.com/LiuHC0428/LAW-GPT}  \\
    
    \hline
    LaWGPT & Chinese-LLaMA, ChatGLM & \url{https://github.com/pengxiao-song/LaWGPT} \\
    
    \hline
    LexiLaw  & ChatGLM-6B & \url{https://github.com/CSHaitao/LexiLaw}  \\
    
    \hline
    Lawyer LLaMA  & LLaMA & \url{https://github.com/AndrewZhe/lawyer-llama}  \\
    
    \hline
    HanFei  & BLOOMZ-7B1 & \url{https://github.com/siat-nlp/HanFei}  \\
    
    \hline
    ChatLaw  & Jiangzi-13B, Anima-33B & \url{https://github.com/PKU-YuanGroup/ChatLaw}  \\
    
    \hline 
    Lychee  & GLM-10B & \url{https://github.com/davidpig/lychee_law}  \\
    
    \hline
    JurisLMs  & AI Judge, AI Lawyer & \url{https://github.com/seudl/JurisLMs}  \\
    
    \hline
    Fuzi.mingcha  & ChatGLM & \url{https://github.com/irlab-sdu/fuzi.mingcha}  \\
    
    \hline
    \end{tabularx}
\end{table*}

\subsection{Fine-tuned Models}

\textbf{LawGPT\_zh} is an open-source Chinese legal language model based on ChatGLM-6B LoRA 16-bit instruction fine-tuning. It incorporates legal question-and-answer datasets and is guided by high-quality legal text question-and-answer construction datasets built from legal articles and practical case studies. This model enhances the performance, reliability, and professionalism of general language models in the legal domain.

\textbf{LaWGPT} \cite{nguyen2023brief} is a series of models that expand the scope of legal terminology and perform pre-training on large-scale Chinese legal text databases to enhance the basic semantic understanding abilities of large models in the legal domain. The models are also fine-tuned on legal dialogue question-and-answer datasets and judicial examination datasets to strengthen their understanding and execution abilities in legal contexts.

\textbf{LexiLaw} is a Chinese legal language model that is fine-tuned based on the ChatGLM-6B architecture. It enhances its performance and professionalism in providing legal consultation and support through fine-tuning legal domain datasets. LexiLaw aims to provide accurate and reliable legal consultation services for legal professionals, students, and general users, addressing specific legal issues, legal articles, case analysis, and legal interpretations, and offering helpful recommendations and guidance.

\textbf{Lawyer LLaMA} \cite{huang2023lawyer,touvron2023llama} is a Chinese legal LLM trained on a large-scale legal dataset. It can provide legal advice, analyze legal cases, and generate legal articles.

\textbf{HanFei}: A fully parameterized Chinese legal LLM with 700 million parameters. It offers functions such as legal question-answering, multi-turn dialogue, article generation, and search.

\textbf{ChatLaw}: A series of open-source legal LLMs developed by Beijing University \cite{cui2023chatlaw}. It includes models such as ChatLaw-13B and ChatLaw-33B, which are trained on a large dataset of legal news, forums, and judicial interpretations. ChatLaw-Text2Vec uses a dataset of 930,000 court cases to train a similarity-matching model.

\textbf{Lychee}: A fine-tuned Chinese legal LLM based on the Law-GLM-10B architecture. It offers better performance and professionalism in legal consultation and support.

\textbf{WisdomInterrogatory}: A legal LLM developed by Zhejiang University, Alibaba, and Hua Research. It uses a pre-training model and domain-specific data to perform legal question-answering and reasoning.

\textbf{JurisLMs}: A collection of LLMs trained on Chinese legal datasets. AI Judge is an explainable legal judgment prediction model that combines a GPT2 model with a legal applicability model. AI Lawyer is an intelligent legal consultation model that can answer questions and provide relevant legal regulations.

\textbf{Fuzi.mingcha}: A model based on the ChatGLM architecture, trained on a large dataset of unsupervised Chinese legal texts and supervised legal fine-tuning data. It supports functions such as legal article search, case analysis, deductive reasoning, and legal dialogue.

\subsection{Evaluation Indicators for the AI and law system}

\begin{figure*}[ht]
    \centering
    \includegraphics[scale=0.4]{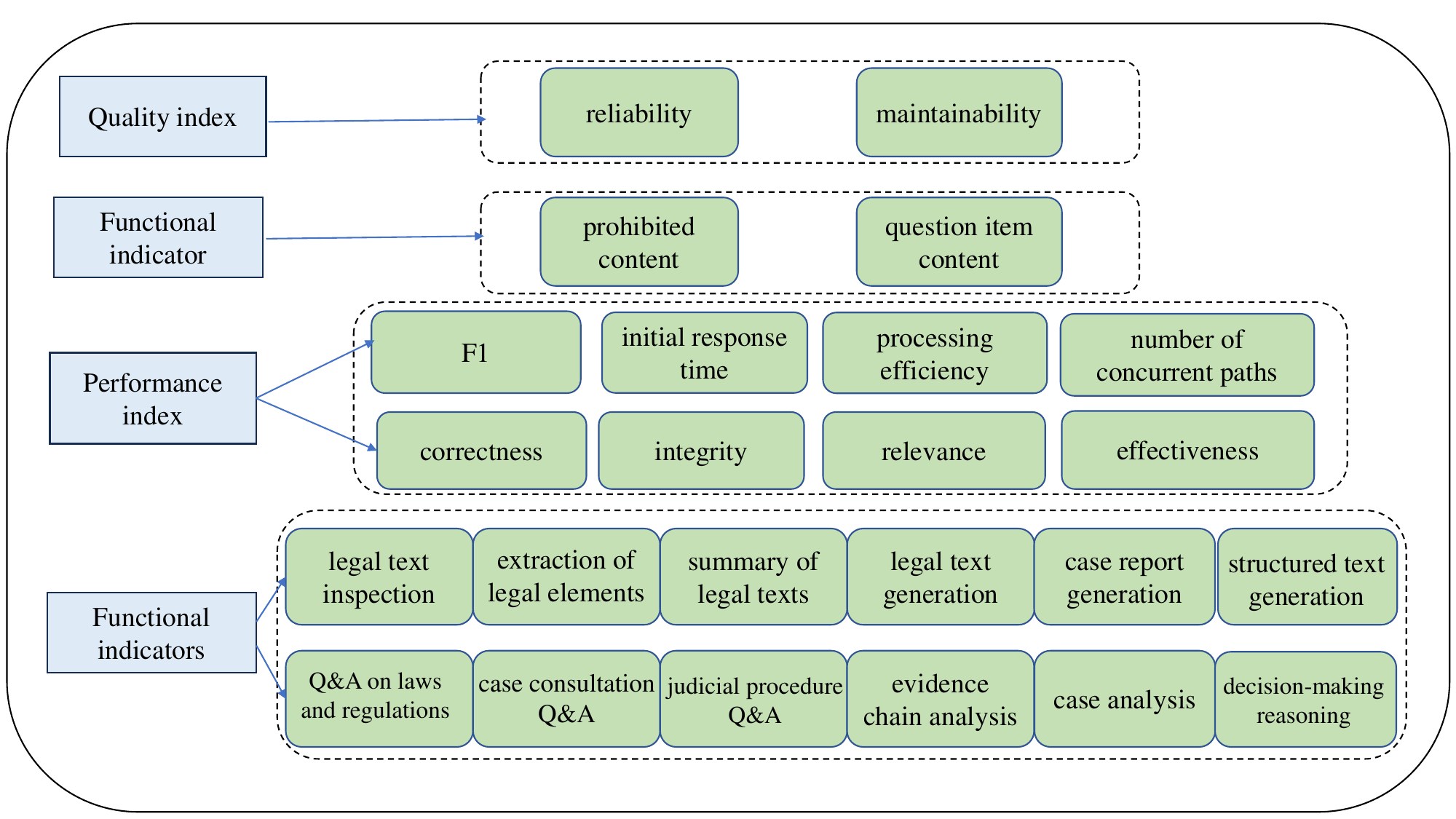}
    \caption{There are various indicators for the AI and law system.}
    \label{fig:assess}
\end{figure*}

In August 2023, a joint proposal was released by several research institutions and universities. The proposal suggests a ``pretrain-prompt-predict" learning model, which utilizes prompt words to fine-tune pre-training models and assess their performance. The proposal suggests a comprehensive evaluation system that combines subjective and objective indicators. The subjective indicators are evaluated by legal experts, while the objective indicators are weighted and scored through a combination of weights and ratings. The evaluation system for legal LLMs consists of two levels, with the first level including four primary indicators: functional indicators, performance indicators, safety indicators, and quality indicators, as shown in Fig. \ref{fig:assess}. The second level of evaluation indicators further breaks down each primary indicator into more detailed sub-indicators.  

\subsection{The Case of Law+AI}

With the increasing application of AI in the judicial field, judicial AI systems are gradually becoming popular. Although the current use of judicial AI is still in its early stages, various measures have been taken in recent years to promote the integration of the judicial industry and AI. For example, the ``206" criminal case assistance system in Shanghai \cite{cui2020artificial}, the ``Mobile Micro-Court" in Zhejiang \cite{shi2021smart}, the Hangzhou Internet Court's smart judgment system \cite{sung2020can}, and the ChatLaw legal LLM released by Beijing University in 2023 are all examples of the application of AI in the judicial field. In other countries, AI is more commonly used to assist judges in making decisions and assessing the risk of criminal behavior. For example, in 2023, a court in Colombia used the ChatGPT decision-making result to assist judges in sentencing a medical insurance case \cite{aydin2023chatgpt}. The HART intelligent system in the UK assists judges in criminal convictions \cite{greenstein2022preserving}, and the ProKid 12-SI system in the Netherlands assesses the possibility of recidivism \cite{la2020legal}. In addition, some countries are committed to researching how to better use AI in handling small cases \cite{aini2020summary,nowotko2021ai}, allowing human judges to focus on more complex cases. This not only improves the efficiency of the judicial system but also helps to reduce the workload of human judges and improve the accuracy of judgments. As AI technology continues to advance, it is expected to play an increasingly important role in the judicial field, revolutionizing the way justice is served.

\section{Challenges} \label{sec:challenges}

Legal LLMs still face many challenges, including the defects of the datasets and shortcomings in the model algorithms. The impact on the traditional legal industry and issues that arise in specific judicial practices are also significant challenges for legal LLMs. Let's introduce more knowledge about this aspect.

\subsection{Defects in Datasets}

The development level of judicial AI is determined by the legal dataset used for training, especially the annotated dataset. Regarding the dataset, judicial AI mainly involves obtaining more representative, comprehensive, and high-quality annotated datasets. This includes addressing unfair elements in the dataset, potential privacy leaks, ensuring the timeliness of legal datasets, and emphasizing the scalability of the dataset.

\textbf{Data inadequate acquisition}: The success of legal LLMs depends heavily on data. Legal big data is characterized by variability and non-structured data. Therefore, the construction of successful legal LLMs relies on the acquisition, organization, and deep learning of legal big data. Given this, the inadequate acquisition of legal data poses a challenge to legal large models. 

\begin{enumerate}[i)]
    \item Insufficient sources of judicial data and documents: In the field of AI and law, the current large-scale legal datasets are insufficient \cite{chalkidis2019deep}. To prevent the leakage of legal data, courts tend to be conservative in terms of the openness of judicial data and legal documents, resulting in a limited variety of legal procedural documents. Consequently, deep learning by legal LLMs cannot cover all legal datasets, which may lead to judicial errors such as misjudgment in practical judicial applications.

    \item Insufficient sharing of legal data \cite{tenopir2011data}: Moreover, due to inconsistent sharing permissions for legal big data among different levels of courts, the legal data used by LLMs is not standardized. Generally, higher-level courts have stricter sharing permissions compared to their subordinate branches, and lower-level courts cannot access data from higher-level courts. Moreover, constrained by sharing technologies, difficulties in data integration, incompleteness, inaccuracies, and outdated legal data can lead to decision-making errors in legal LLMs.

    \item Non-standard legal documents: Furthermore, due to the lack of standardized review or requirements for legal documents in some courts and the insufficient responsibility of some legal professionals, some legal documents do not meet the standards. Inaccurate and non-standardized legal documents are difficult for algorithms within legal LLMs to recognize, resulting in incomplete data acquisition.
\end{enumerate}

\textbf{Inaccurate interpretation of legal concepts}:
Legal concepts can be classified into evaluative concepts, descriptive concepts, and discursive concepts \cite{hage2009concepts}. However, legal concepts have inherent uncertainty in their connotations and extensions. Current AI systems have recognition deficiencies when dealing with legal concepts, which can lead to inappropriate assumptions and deficiencies in the identification of concepts. For example, big data may derive unknown conclusions or misinterpret the boundaries of legal concepts. Judges typically consider the specific context and value assessment when applying specific legal concepts, which are challenging for legal LLMs to achieve. Recognizing and structuring fuzzy boundary concepts in law poses a significant challenge to the development of legal LLMs.

\textbf{Characteristics of the datasets}: 
\begin{enumerate}[i)]
    \item \textit{Timeliness}: In the legal system, the specific application and meaning of legal concepts may evolve and improve over time depending on the location. When the judicial datasets used for training legal LLMs do not include these subtle changes, the results generated by AI may not be suitable for the latest legal environment, leading to judicial misjudgments. 

    \item \textit{Credibility}: Due to the variety and complexity of laws, the large number of judicial cases and documents, and the complexity of the training process for legal datasets, biased elements in data acquisition, inadequate selection of inconsistent, incomplete, or erroneous data, improper privacy protection in the datasets \cite{de2012data}, and imbalanced training of the datasets may exist. For example, civil litigation cases may be more common than criminal cases or other types, resulting in imbalanced training datasets and poor performance of judicial AI in certain cases. 

    \item \textit{Scalability}: Currently, the scale of the datasets is limited, and some legal datasets only include cases from specific time periods or the application of individual legal articles at specific levels, making it difficult to extend to other aspects.
\end{enumerate}

\subsection{Shortcomings in Algorithms}

LLMs in the field of judiciary have garnered significant attention regarding algorithmic interpretability, ethics, bias and fairness, and algorithmic optimizability. In this subsection, we delve into the shortcomings of LLMs in these aspects.

\textbf{Interpretability}: The use of deep learning, neural networks, and other algorithms in LLMs makes their structures complex, and the results of their decisions are difficult to predict. Current LLMs cannot achieve complete value neutrality, and the authority of the judiciary is based to some extent on the public's acceptance of the outcome of case handling. Insufficient interpretability of large models undoubtedly reduces people's trust in the judicial application of AI \cite{atkinson2020explanation}. Therefore, interpretability is a key challenge for legal LLMs, and it is crucial to reduce the black-box nature of legal LLMs and improve their interpretability.

\textbf{Ethics, bias, and fairness}: The fairness and security of algorithms behind legal LLMs have received significant attention \cite{wachter2021fairness}. 

\begin{enumerate}[i)]
\item Algorithms may contain elements of inequality. Some data may contain elements of gender or racial discrimination, which may also be influenced by historical factors. For example, the COMPAS algorithm made different prediction errors for white and black individuals \cite{flores2016false}. Black individuals were erroneously predicted as high-risk, while white individuals were erroneously predicted as low-risk. Such algorithmic discrimination may exacerbate racial bias. The National Institute of Standards and Technology categorizes these biases into statistical bias, human bias, and systemic bias \cite{schwartz2022towards}. However, when AI learns from this data, it cannot judge the fairness of the data and therefore cannot eliminate unfair factors. In practical judicial applications, these unequal elements may unintentionally be amplified by LLMs, resulting in unfair decisions.

\item Insufficient security in algorithm outsourcing: Moreover, due to a lack of legal-technical hybrid talent, the algorithmic part of legal LLMs is often outsourced to algorithm companies, which unintentionally reduces transparency in legal interpretations. The model algorithms may be exposed to ``biased" elements or elements that do not align with human values, which are then ``technically cleansed" by AI, hiding these unjust elements. This can lead to issues such as ``algorithmic black boxes" \cite{yu2019s}. For example, when an AI system provides a judgment in a significant case, the party being judged may find it difficult to be convinced by the decision if the system cannot provide a reasonable explanation for the underlying reasons or algorithmic principles behind the ruling.

\item  Moreover, the reduced transparency of LLMs in law may lead to significant issues of judicial unfairness \cite{rai2020explainable}, and trust in the judiciary may decrease. During the development stage of legal LLMs, the handling of most legal concepts and unstructured data such as legal documents, including collection, cleaning, annotation, and processing, may introduce elements of inequality. For example, if AI does not have a comprehensive understanding of sensitive information (such as race), it may be challenging to completely remove sensitive elements during the cleaning process of legal datasets, incorporating biased elements into the decision-making model. During the testing stage of the model, developers may make adjustments to the model's decision results based on their own cultural background and scope of understanding. However, developers may not be legal professionals, resulting in biases in the understanding of legal concepts and specific judicial discretion. This can lead to ``blind spot bias", such as the issue of racial bias in the Gangs Matrix's prediction of potential offenders \cite{densley2020matrix}.

\item If the algorithmic bias or lack of interpretability behind legal LLMs cannot be improved and becomes a significant cause of AI judicial errors, the public will distrust the judicial application of legal LLMs, thereby constraining the development of legal LLMs.
\end{enumerate}

\subsection{Challenges of Traditional Legal Industry} 

The rapid development of the digital age and AI has led to a shift towards proactive justice, creating a challenge for the traditional legal industry known as ``activist justice". The specific challenges are as follows:

\textbf{Neglecting Judicial Independence}: Currently, legal LLMs are not sufficient to replace judges in making decisions \cite{zavrvsnik2020criminal}. Judicial independence encompasses aspects of legal enforcement and fact-finding. In terms of legal enforcement, independence includes interpreting civil law, explaining uncertain concepts, and evaluating disputes over the rights and interests of parties involved in a case. In terms of fact-finding, independence includes the use of discretion, subjective judgment, experiential judgment, and weighing pros and cons in decision-making. For example, judges need to make judgments based on the evidence and testimony presented by both parties during litigation. In these aspects of judicial independence, legal LLMs often play an auxiliary role. However, if legal LLMs excessively intervene in case decisions, it can lead to judges overly relying on AI to find relevant literature, establish facts, and even form preconceived notions before the proceedings, thereby neglecting the demands of the parties involved in the case. 

This weakens the exercise of judicial discretion. The decision-making solutions generated by legal LLMs based on case characteristics may deprive judges of their discretionary powers in the details of a case \cite{fagan2019impact}. Judges have the freedom to exercise discretion in aspects such as assessing the credibility of evidence based on the applicability of the law. They can also make reasonable decisions by considering factors such as the extent of the victim's property loss and the compensation ability of the defendant, drawing from judicial experience or from the perspective of the parties involved. For example, in assessing the compensation amount in civil litigation, judges can comprehensively consider factors such as the extent of the victim's financial loss and the defendant's ability to compensate. In contrast, the algorithms of legal LLMs struggle to measure the extent of loss and evaluate a person's ability to pay, so decisions based solely on case characteristics to a certain extent weaken the judge's discretionary powers.

The positioning of legal LLMs is unclear. The unclear positioning of legal LLMs undermines the role and judicial power of judges in cases. Legal LLMs should assist judges in decision-making \cite{rissland2003ai}, provide advice, and automatically generate legal texts, etc., but it does not possess professional judicial experience and cannot independently make judgments in cases. Therefore, users should fully understand the position of legal LLMs within the legal system when utilizing them.

\textbf{Impact on the traditional court system}: The concept of trial centrism emphasizes the principle of equality between prosecution and defense, with the judge as the main authority and the trial playing a decisive role. With the development of AI technologies such as AIGC, legal LLMs have alleviated the problem of manpower shortages in the legal field but have also restrained the subjective initiative of judges and the development of traditional trial systems. This is mainly reflected in the following aspects:

Court idleness: Trial centrism emphasizes the central role of judges in the judicial process \cite{re2019developing}, and a fair trial allows equal confrontation between the parties involved, enabling them to have confidence in the judicial process and reach a just judgment. However, the popularization of legal AI systems may lead to court idleness, which would diminish the solemnity of the legal process and the subtle influence it brings. The idleness or weakening of court trials will restrict the subjective initiative of judges in the judicial process and reduce public trust in the judiciary.

Crisis in the hierarchy of trials: The hierarchical system establishes a relationship between higher and lower courts, ensuring people's litigation rights and obtaining fair and prudent judicial outcomes by setting different levels of adjudication. Legal AI systems and other AI technologies may impact the judicial process in the hierarchical system \cite{contini2020artificial}. For example, if a party is dissatisfied with the judgment of a lower court and appeals to a higher court for a second trial. If most courts use the same legal AI system, the second trial would be no different from the first trial. This goes against the purpose of safeguarding litigation rights, the supervision of lower courts by higher courts, and the achievement of judicial fairness through the hierarchical system. Therefore, the application of legal LLMs in the judicial field may have certain impacts on the trial system.

\subsection{Issues Arising from Specific Judicial Practice}

\textbf{The lack of universality in applications.} Legal LLMs, when assisting in decision-making \cite{zhong2020does}, often extract feature values from cases and search for similar cases within existing multidimensional datasets to find the ``optimal solution". However, due to the differences between cases, the ``optimal solution" proposed by the large model may not be applicable to a particular case. Furthermore, legal regulations may vary across different countries or regions, leading to inconsistent decision outcomes for the same case under different legal rules. Legal LLMs struggle to address the issue of case diversity and cannot be applied to all legal cases.

\textbf{The lack of subjective thinking, emotions, and experience.} Compared to legal professionals, legal LLMs lack autonomous thinking abilities and professional experience, among other things. Legal LLMs can process cases through factor identification, but judicial experience is difficult to quantify accurately (e.g., the standard of proof for ``reasonable doubt" in criminal proceedings), and AI struggles to subjectively assess the truthfulness of case statements. Moreover, legal AI systems lack the elements of blending law and empathy, which undoubtedly results in a lack of human touch in legal regulations and undermines people's trust in the judiciary. Simultaneously, the judicial decision-making process is not merely a logical reasoning process on a single layer but also involves moral, ethical, and practical considerations in the legal system \cite{xu2022human}. Given these limitations, there are still deficiencies in the application of legal LLMs in the field of justice.

\textbf{Contradiction with the presumption of innocence principle.} In recent years, AI systems have been applied in preventive crime measures, exemplified by the COMPAS system for crime prediction and risk assessment \cite{beriain2018does}, PredPol for iterative calculation of potential crime locations through the analysis of criminal history data \cite{rosser2017predictive}, and the PRECOBS system in Germany for burglary prevention and violence crime prediction \cite{egbert2020predictive}. This has led to a shift in policing from being 'service-oriented' and 'reactive' to 'proactive prediction' \cite{hardyns2018predictive}. However, this shift actually contradicts the principle of presumption of innocence. Judicial decisions should be based on known cases, and predictive modeling algorithms based on personal privacy data may deprive individuals identified as ``future potential criminals" of basic public services such as education or social welfare, leading to discrimination and restrictions. However, these future-oriented ``judgments" fundamentally undermine human rights. Moreover, as predictive policing becomes more prevalent and the number of registered ``tagged offenders" increases, a particular area may be perceived as a ``high-crime area" due to these model-driven decisions, inadvertently promoting regional discrimination and bias. This contradicts both human rights and the presumption of innocence principle.

\begin{enumerate}[i)]
    \item Imbalance of prosecution and defense. The application of AI technologies, such as legal LLMs, in the field of the judiciary may lead to an imbalance of public and private powers. The application of AI in the judiciary will result in the problem of ``triangular imbalance", causing an imbalance in the relationship between prosecution and defense, and the excessive exercise of public power is accompanied by a decrease in public trust.

    \item Unequal control over data. Controlling over the big data used by LLMs is in the hands of public authorities or large corporations, making it difficult for individuals to access the operational mode of big data \cite{prescott2017improving,zuiderveen2020strengthening}. This is likely to lead to the expansion of public power and the contraction of private rights. For example, during the investigation process, public authorities can access case data and collect criminal evidence about suspects through methods such as software or mobile phone tracking. However, the defense side, due to its own power limitations, is in a disadvantaged position when collecting more relevant data. The unequal control over data greatly restricts the fairness of the trial.

    \item Differences in the ability to analyze case data. Some research \cite{wexler2021privacy} believes that even if both the prosecution and defense have the same access rights to obtain data, there is still a significant gap in their ability to analyze case data \cite{wexler2021privacy}. The prosecution can extract and analyze legal case data based on national regulations, utilizing national resources such as a large number of professional data analysts and advanced data analysis equipment. Most defense lawyers obviously lack this kind of professional data analysis capacity, and a large amount of data analysis increases litigation costs for citizens.

    \item Inconsistent attention to legal LLMs. The application of legal LLMs reflects issues of policy attention, investment imbalance, and unequal exploration results between public power departments such as public security agencies, judicial organs, and individuals. This will reduce the public's trust in the judiciary.

    \item Administrative intervention leads to the abuse of legal LLMs. Individual performance assessment is an important evaluation criterion for judges. With the gradual application of legal LLMs, some courts incorporate indicators of AI-assisted decision-making into the evaluation criteria for judges. This, to some extent, has resulted in the abuse of AI in the field of the judiciary and has also undermined the status of judges in the trial process.
\end{enumerate}

\textbf{Privacy infringement.} In practical applications, legal LLMs are prone to privacy infringement issues \cite{rodrigues2020legal,zavrvsnik2020criminal}. Firstly, legal LLMs may collect excessive user data, which can lead to privacy breaches when the input information is sensitive. Secondly, the underlying algorithms of legal LLMs may be inadequate, resulting in improper data handling. For example, if user inputs contain sensitive labels (such as birthdates or home addresses) in the legal dataset used for deep learning and the model lacks proper anonymization procedures, user privacy could be amplified, leading to privacy infringement issues.

\textbf{Issues of intellectual property identification and protection.} With the continuous development of AI technology, the identification and protection of intellectual property rights in the legal field have attracted attention \cite{rodrigues2020legal}. The involvement of AI in creative processes makes it difficult to determine the ownership of intellectual property. It becomes unclear whether credit should be given to the human users who utilize AI or to the AI system itself. Since AI lacks independent thinking capabilities and its creations are derived from deep learning on existing works within the field, it may not meet the requirement of originality. Therefore, it can be inferred that legal LLMs and other AI technologies may amplify the deficiencies in the identification and protection of intellectual property rights in the judicial domain.

\subsection{Ethical Views Impacting Human Society}

\textbf{Disregard for human subjectivity.} During the training process of LLMs on datasets, in order to ensure the values of the LLMs themselves, negative comments in the dataset need to be labeled and filtered. However, due to the low cost of labor, some workers who are hired at very low prices may experience psychological problems when screening a large number of negative comments, and human subjectivity is susceptible to algorithmic bullying. In addition, the labor contribution of humans in this process is also being ignored. Legal LLMs are challenging the ethical views of human society \cite{raso2018artificial}.

\textbf{Misleading user comments.} In testing certain LLMs, such as ChatGPT, AI has displayed behaviors such as inducing users to divorce, making inappropriate comments, and even encouraging users to disclose personal privacy or engage in illegal activities \cite{riveiro2021s}. The reason is that the training dataset of AI contains improper or discriminatory comments, and through deep learning, AI acquires these characteristics. This undoubtedly has an impact on the ethics of human society.

\textbf{Ethical value consistency.} The development of LLMs should first and foremost adhere to the ethical values of all humanity. If the values of AI cannot align with those of humans, there may be situations where AI misleads or harms human interests, which poses challenges to national governance systems and global cooperation.

\section{Future Directions} \label{sec:future} 

The rapid development of legal LLMs is continuously changing the landscape and practices in the legal field. With the ongoing advancements in AI and natural language processing technologies, we are paying attention to more issues related to the applications of legal LLMs, and we recognize that they hold broader prospects for development. We propose several potential directions for the future development of legal LLMs. Details are discussed below. 

\subsection{Data and Infrastructure}

\textbf{Obtaining more comprehensive legal big data}: Firstly, we should broaden the scope of obtaining legal big data \cite{zhong2020jec}. For example, we can collect data not only from specific court databases but also enhance the protection techniques for legal data, such as encrypting data transmission and increasing access control, in order to promote greater openness of data by various courts. Secondly, to address the issue of insufficient sharing permissions, we can improve legal regulations and clearly define the sharing permissions between different levels of courts, stipulating their rights and obligations, thus facilitating more sharing of legal data. Leveraging cloud computing to enhance new sharing technologies and establishing an open and secure big data sharing platform. Lastly, courts can standardize legal document formats and conduct preprocessing operations, such as evaluating and cleaning large amounts of legal data using unified filtering criteria. Moreover, optimizing pre-training models (e.g., Lawformer) for legal documents can enhance training on long legal documents \cite{xiao2021lawformer}.

\textbf{Defining the boundaries of legal concepts and limiting the scope of application}: To address the challenge of legal LLMs struggling to convert certain ambiguous legal concepts into programming symbols, we have summarized some solutions. Firstly, it involves defining the boundaries of legal concepts and restricting fuzzy legal concepts based on criteria like societal impact \cite{rissland2003ai}. Secondly, it entails limiting the applicability of legal LLMs in legal cases and reducing their decision-making proportion in cases involving relevant legal concepts while upholding the judicial authority of judges.

\textbf{Data transparency}: Whether it is ChatLaw or Legal BERT, both legal LLMs require a series of operations such as collecting, screening, extracting, classifying, and training legal datasets. These processes may introduce bias or adopt unreasonable and incomplete methods. Therefore, it is necessary to establish and appropriately disclose standards for datasets and AI mechanisms, including dataset content, usage restrictions, licenses, data collection methods, information on data quality and uncertainty, as well as data features, structure, and classification schemes \cite{zhang2023intelligent}. Openly sharing data promotes more comprehensive and accurate datasets for large-scale model training, while also enhancing public trust in the judiciary.

\textbf{Building a legal knowledge graph}: This enables the connection of legal concepts, legal cases, and applicable legal rules, establishing a model that is easily understandable by AI. Firstly, legal concepts such as ``criminal law" and ``intellectual property rights" are classified. Then, logical relationships are established for applicable legal rules, such as ``criteria for determining illegality". Specific legal cases, such as ``robbery cases", are classified. Finally, by referencing legal documents in the legal knowledge base, a comprehensive legal knowledge graph is established \cite{yang2021legalgnn}, thereby optimizing the functionality of the legal model.

\textbf{Optimizing the foundational infrastructure for model training}

\begin{enumerate}[i)]
\item High-performance computing resources: Considering the vast variety and quantity of legal datasets, it is recommended to utilize high-performance computing resources such as Graphics Processing Units (GPUs) \cite{brodtkorb2013graphics} and Tensor Processing Units (TPUs) \cite{jouppi2017datacenter} to significantly improve the training and inference speed of legal LLMs \cite{scao2022language}.

\item Distributed computing frameworks: Single-model training is no longer sufficient to meet specific requirements. Consider using distributed computing frameworks, such as TensorFlow \cite{abadi2016tensorflow} and PyTorch's distributed training capabilities \cite{paszke2019pytorch}, to achieve parallel processing and accelerate model training \cite{li2020pytorch}. By distributing computing tasks across multiple nodes and devices, the training process can be completed more quickly.

\item Storage and data management: Appropriate storage and data management solutions need to be considered for handling datasets of legal LLMs. For example, using high-performance storage systems like SSDs (Solid State Drives) can provide fast data read and write speeds \cite{chen2011essential}. Moreover, distributed file systems or object storage systems can be used to effectively manage large-scale legal datasets.

\item Data preprocessing and cleaning: Proper preprocessing and cleaning of data are necessary before using legal LLMs. This includes removing noise and redundant information, standardizing text formats, and ensuring the quality and consistency of the dataset. Utilizing professional data cleaning tools and techniques can help improve the performance and accuracy of legal LLMs.

\item Model scaling and deployment: When deploying legal LLMs into practical applications, considerations must be given to model scaling and deployment issues. Model compression and pruning techniques can reduce the size and computational requirements of the model, improving deployment efficiency on edge devices. Moreover, containerization technologies such as Docker and Kubernetes \cite{bernstein2014containers} can be used to simplify model deployment and management \cite{kozhirbayev2017performance}.
\end{enumerate} 

\subsection{Algorithm Level}
\textbf{Strategy adjustment and optimized algorithms}: To address the challenges of modeling long texts, we can employ external recall methods, model optimization \cite{hoffmann2022training}, and optimization of the attention mechanism. These approaches can help us tackle the difficulties of long text modeling and improve the capability and effectiveness of legal LLMs in handling long texts. External recall methods involve utilizing external tools or external memory to assist in processing long texts. In the legal domain, a legal knowledge base can be established, containing legal documents, precedents, regulations, and other information. The model can query this knowledge base to obtain relevant legal background and explanations, enabling better understanding and processing of long texts. Model optimization involves improving the modeling capability for long texts by optimizing the model itself. In the legal domain, specialized legal models can be designed and trained to address specific tasks and requirements. These models can be trained using legal domain data during the pre-training phase to enhance their understanding and processing capabilities for legal texts. Optimization of the attention mechanism, which is a critical component of large models, can improve the effectiveness of long-text modeling \cite{niu2021review}. In the legal domain, more complex and refined Attention mechanisms can be designed to enable the model to better focus on key information and context within the text.

\textbf{Limiting algorithmic biases and ``black box" operations}: When elements of inequality, such as racial or gender biases, are incorporated into the algorithmic decision-making of LLMs, it can lead to different legal judgments in criminal law, resulting in unjust judicial outcomes. Moreover, the opacity of certain legal LLMs makes it difficult to prove the absence of ``black box" operations \cite{pedreschi2019meaningful}. Making algorithms public and subjecting them to evaluation can enhance the transparency and openness of legal LLMs, thereby facilitating their use in assisting judicial decision-making \cite{liang2021towards}.

\begin{enumerate}[i)]
\item Implement protective algorithms: In the process of handling judicial data at the underlying algorithmic level of LLMs, algorithms can be introduced to protect against biased elements. For example, random noise can be added to the original data, and algorithms based on differential privacy can be applied to protect the data \cite{wei2020federated}.

\item Implement algorithmic explainability: With the application of explainable artificial intelligence (XAI) \cite{das2020opportunities,dovsilovic2018explainable}, we need to consider how to better integrate XAI and legal requirements, and how to further enhance the ability of XAI algorithms to identify biases \cite{deeks2019judicial}. By increasing the explainability of AI decision-making algorithms, we can promote the transparency of LLM judgments and restrict the occurrence of unfair operations.

\item Incorporate legal indicators for evaluation: Solaiman \cite{solaiman2021process} proposes a Process for Adapting Language Models to Society (PALMS) with Values-Targeted Datasets \cite{solaiman2021process}. The method is designed to ensure that the models are fair and do not discriminate against certain groups. The evaluations are based on quantitative metrics such as similarity between output and target results, toxicity scoring, and qualitative metrics that analyze the most common words associated with a given social category. By using legal metrics to evaluate the decision-making results of the LLMs, the method can effectively address potential bias issues. Additionally, some countries have enacted laws to prevent algorithmic bias, such as the European Union's laws protecting the rights of individuals against algorithmic discrimination, including data protection laws and non-discrimination laws \cite{zuiderveen2020strengthening}.
\end{enumerate} 

\textbf{Promote limited algorithmic transparency}: Making decision algorithms public allows people to gain a better understanding of the underlying principles, whether it is the weighing of judgments in common civil litigation or the criminal discretion of public authorities. Even small algorithmic changes or the definition of a biased element can introduce difficult-to-assess risks. Decision-making bodies should focus on promoting the transparency of decision algorithms, allowing legal LLMs' algorithms to be supervised by all parties involved, promoting judicial transparency, and achieving fairer judgments. For example, in the case of Wisconsin v. Loomis \cite{beriain2018does}, the refusal to disclose the judgment results based on trade secrets clearly undermines public trust. However, algorithmic transparency should be subject to certain constraints. For instance, in cases involving potential trade secrets or national security information, flexibility should be allowed. Limited algorithmic transparency can be achieved through the signing of confidentiality agreements with parties involved or closed pretrial hearings. In addition, algorithmic transparency enables ordinary citizens with legal consulting needs to input keywords that better match the algorithmic procedures, thereby obtaining legal advice that is more tailored to the characteristics of their cases.

\subsection{Dealing with Traditional Judiciary}

\textbf{Clarifying the positioning of large models}: legal LLMs often intervene in judicial decision-making and even influence the discretionary power of judges. This is mainly due to the unclear positioning of legal models within the legal system \cite{re2019developing}. To address this issue, we should uphold the independence and autonomy of judges and improve relevant laws and regulations to clearly define the role, functions, and scope of legal AI in the judiciary. We should utilize large models to assist judges in decision-making, such as providing precise references to legal documents, rather than relying solely on LLMs for decision-making. At the same time, with the assistance of AI, judges should strive to become more empathetic, rational, and professional decision-makers.

\textbf{Defining the thinking capability of LLMs}: Legal LLMs process a vast amount of legal datasets through algorithms like deep learning to simulate judges' information extraction and decision-making in cases. LLMs do not, however, have independent thinking abilities due to computational limitations. Therefore, we need to define the thinking capability of large models, and here are some evaluation indicators. Firstly, we can assess the predictive accuracy of legal LLMs by comparing the predictions for a certain type of case with the outcomes of traditional judicial systems \cite{mckay2020predicting}. Secondly, we can evaluate the extent of resource consumption. For example, we can compare and evaluate the time, resource, and manpower consumption of processing a given case using a legal LLM versus traditional judicial methods. Furthermore, explanations should be provided by the legal LLMs regarding the connection between the case and the judgment results, and these explanations can be compared with those provided by judges to calculate the degree of alignment between AI and human judicial thinking.

\textbf{Ensuring parties' access to data}: Parties involved in judicial cases are the main subjects of the cases and the primary sources of data used by AI. They should have the right to access, question, and update their data \cite{zuiderveen2020strengthening}. Scholars argue that ``data access rights" are essential in a technology-based judiciary \cite{alzain2011using,prescott2017improving}. In the process of AI-assisted judicial decision-making, the parties' legal access rights should be protected. For example, they should have the right to question the digitized evidence of their case and the right to update outdated data used by AI. In addition, the defense should have the right to contact technical personnel related to AI, ensuring their ability to review and question the data used by legal LLMs.

\textbf{Expanding and optimizing the consulting function of judicial large models}: AI technologies such as LLM and ChatGPT have interactive question-and-answer capabilities. In the field of judicial judgments, expanding the consulting capabilities of AI may provide more opportunities for individuals who are not familiar with the law to receive legal advice. Optimizing the legal consulting capabilities of the LLM involves specific applications and constraints of criminal law, for example.

\begin{enumerate}[i)]
\item Fine-tuning the LLMs and introducing multimodal capabilities. The model can be fine-tuned for different subfields of law. For example, to optimize the consulting capabilities in the field of labor law, the model can be fine-tuned using labor law-related cases and materials to enhance its performance in that subfield. Additionally, to better understand user needs, multimodal inputs can be introduced \cite{joshi2021review}, such as allowing users to provide legal questions through voice or images. Similarly, the model can provide multimodal outputs based on user needs and scenarios, such as speech, text, charts, etc.

\item Involvement of legal experts and practitioners. Inviting legal experts, judges, prosecutors, and lawyers to participate in the development and evaluation process of judicial AI. They can provide valuable legal knowledge and practical experience to help guide the design and training of AI systems, ensuring compliance with legal regulations and judicial practices. Furthermore, legal experts can evaluate and refine the results of the LLMs \cite{wachter2021fairness}, fine-tuning it to improve the accuracy and reliability of the model in legal consulting.

\item User feedback and continuous optimization: In the training process of the legal LLMs, the Reinforcement Learning from Human Feedback (RLHF) machine learning method can be adopted \cite{liu2023summary}. First, provide users with a feedback mechanism \cite{ouyang2022training}, encourage them to actively participate in the use of legal LLMs, and collect user feedback. According to the annotated data containing user feedback, train a reward model (RM) to rank the content generated by the large language model. Finally, update the parameters of the LLMs through deep learning, making the legal LLMs more in line with the actual needs of users. This method reduces the problem of needing extensive trial and error in traditional reinforcement learning and optimizes and adjusts the model based on user needs.
\end{enumerate} 

\subsection{Judicial Practice}

\textbf{Improve accountability mechanisms to prevent political interference}: Establishing a sound accountability mechanism is beneficial for regulating the use of AI in the judiciary and preventing legal errors caused by legal LLMs \cite{contini2020artificial}. For example, the responsibility for judicial errors would discourage judges from completely delegating decision-making to legal LLMs and instead prompt them to carefully consider the decision recommendations provided by legal LLMs. Furthermore, we should guard against political interference in the utilization of AI in the judiciary. For instance, in order to promote the application of AI in the judicial field, courts at all levels may receive performance assessment indicators related to the use of AI technologies. In the pursuit of meeting these indicators, courts may inadvertently overlook the limitations of AI, leading to judicial errors.

\textbf{Foster the development of interdisciplinary talents}: One major reason for issues such as ``algorithmic bias" and ``algorithmic black boxes" in legal LLMs is the lack of interdisciplinary talents combining legal and technological expertise. We should encourage the cultivation of talents that possess both legal and technological knowledge \cite{wei2021innovative}. This would enable us to address the problem of legal concepts that are difficult to translate into algorithmic programs. Moreover, interdisciplinary talents with legal experience would be able to identify discriminatory elements in large datasets and eliminate them. This also ensures the security of algorithms in legal LLMs and greatly reduces the probability of ``algorithmic black boxes" occurring.

\textbf{Collaboration and sharing of experiences}: AI technology has developed rapidly internationally. The research on reasoning capabilities can be traced back to the Mycin system in the 1970s, followed by the establishment of IBM's Watson model, the study of LLMs, and the release of OpenAI's ChatGPT. There is rich experience in the field of AI. It is important to collaborate with foreign research institutions, companies, and experts to share experiences and technological achievements. International cooperation can facilitate exchanges and collaborations among different countries in the field of AI in the judiciary, enabling them to address technical and ethical issues jointly. It can also avoid duplication of effort and errors.

\section{Conclusions}  \label{sec:conclusions}

This paper is dedicated to synthesizing various technologies and ideas regarding the opportunities, challenges, and recommendations for the application of AI in the judicial field. We hope that this paper can give some potential inspiration or research directions to those who are conducting research on legal LLMs or legal practitioners. This paper reviews the opportunities and challenges of legal LLMs in the judicial field. With the rapid development of AI technology, LLMs based on AIGC and other AI technologies have attracted widespread attention in the legal field. We first provide an overview of the development and related research of legal LLMs and explore their opportunities in judge-assisted trials and legal consultation. Next, we discuss the shortcomings of legal LLMs at the algorithmic level and in specific practice. Legal models have huge potential in the judicial field. They can provide efficient legal advice and assist judges in making decisions, speed up the processing of judicial cases, reduce the workload of judges, and improve the accuracy and consistency of judicial decision-making.

However, legal LLMs still face some challenges, such as insufficient processing capabilities for long texts, a limited ability to understand and adapt to individual cases, and privacy and ethical issues. In order to give full play to the advantages of legal LLMs and cope with challenges, we put forward the following suggestions and future development directions: strengthen data quality and privacy protection, improve the long text processing capabilities of the model, improve the adaptability and intelligence level of the model, and establish a reasonable ethical framework and regulatory mechanisms. Future development directions include exploring the application of legal models in other judicial fields, strengthening international cooperation and knowledge sharing, and establishing a cooperation model with multi-party participation to achieve sustainable development and the social benefits of legal LLMs.

\section*{Declaration of Competing Interest}
The authors declare that they have no known competing financial interests or personal relationships that could have appeared to influence the work reported in this paper.

\section*{Acknowledgment}
This research was supported in part by the National Natural Science Foundation of China (No. 62272196), the Natural Science Foundation of Guangdong Province (No. 2022A1515011861), Engineering Research Center of Trustworthy AI, Ministry of Education (Jinan University), and Guangdong Key Laboratory of Data Security and Privacy Preserving.

\printcredits

\bibliographystyle{cas-model2-names}

\bibliography{llmlaw.bib}

\end{document}